%% file: test.tex
\documentclass{article}

\pdfoutput=1
\usepackage[utf8]{inputenc} 
\usepackage[T1]{fontenc}    
\usepackage{hyperref}       
\usepackage{url}            
\usepackage{booktabs}       
\usepackage{amsfonts}       
\usepackage{nicefrac}       
\usepackage{microtype}      

\usepackage{microtype}
\usepackage{graphicx}
\usepackage{subcaption}
\usepackage{booktabs} 
\usepackage{amsmath, amssymb}

\usepackage{amsthm}
\newcommand{\bigCI}{\mathrel{\text{\scalebox{1.07}{$\perp\mkern-10mu\perp$}}}}
\usepackage{hyperref}

\newcommand{\indep}{\rotatebox[origin=c]{90}{$\models$}}

\begin{document}

\title{\textbf{Cause-Effect Deep Information Bottleneck For Systematically Missing Covariates}}

\author{Sonali Parbhoo$^\dagger$, Mario Wieser$^*$, Aleksander Wieczorek$^*$, Volker Roth$^*$}
\date{$\dagger$Harvard John A Paulson School of Engineering and Applied Sciences, Harvard University, USA.\\
  $^*$Department of Mathematics and Informatics, University of Basel, Switzerland\\ \texttt{sparbhoo@seas.harvard.edu}}

\maketitle

\begin{abstract}
\noindent Estimating the causal effects of an intervention from high-dimensional observational data is difficult due to the presence of confounding. The task is often complicated by the fact that we may have a systematic missingness in our data at test time. Our approach uses the information bottleneck to perform a low-dimensional compression of covariates by explicitly considering the relevance of information. Based on the sufficiently reduced covariate, we transfer the relevant information to cases where data is missing at test time, allowing us to reliably and accurately estimate the effects of an intervention, even where data is incomplete. Our results on causal inference benchmarks and a real application for treating sepsis show that our method achieves state-of-the art performance, without sacrificing interpretability.
\end{abstract}

\input{introduction}

\input{foundations}
\input{method}

\input{results}

\input{conclusion}

\small{
    \bibliography{paper}
    \bibliographystyle{plain}
}

\end{document}

%% file: introduction.tex
\section{Introduction}
Understanding the causal effects of an intervention is a key question in many applications, such as healthcare (e.g. \cite{wager2017estimation, AlaaS17}). The problem is especially complicated when we have a complete, high-dimensional set of observational measurements for a group of patients, but an incomplete set of measurements for a potentially larger group of patients for whom we would like to infer treatment effects at test time. For instance, a doctor treating patients with HIV may readily have access to routine measurements such as blood count data for all their patients, but only have the genotype information for some patients at a future test time, as a result of the medical costs associated with genotyping, or resource limitations. 

A naive strategy to address this problem would be to remove all those features that are missing at test time and infer treatment effects on the basis of the reduced space of features. Alternatively, one may attempt imputing the incomplete dimensions for the same purpose. Both of these solutions however, fail in high-dimensional settings, particularly if the missingness is systematic as in this case, or if many dimensions are missing. Other approaches account for incomplete data during training, for instance by assuming hidden confounding. These methods typically try to build a joint model on the basis of noisy representatives of confounders (see for instance \cite{Greenland08, pearl2012measurement, kuroki2014measurement, Louizos}). However, in high-dimensional-settings, it is unclear what these representatives might be, and whether our data meets such assumptions. Regardless of these assumptions, none of these approaches addresses systematic missingness at test time.

A more natural approach would be to assume one could measure everything that is relevant for estimating treatment effects for a subset of the patients, and attempt to transfer this distribution of information to a potentially larger set of test patients. However, this is a challenging task given the high dimensionality of the data that we must condition on. Here, we propose tackling this question from the decision-theoretic perspective of causal inference. The overall idea is to use the Information Bottleneck (IB) criterion \cite{art:tishby:ib, DVIB} to perform a sufficient reduction of the covariates or learn a minimal sufficient statistic for inferring treatment outcomes. Unlike traditional dimensionality reduction techniques, the IB is expressed entirely in terms of information-theoretic quantities and is thus particularly appealing in this context, since it allows us to retain only the information that is relevant for inferring treatment outcomes. Specifically, by conditioning on this reduced covariate, the IB enables us to build a discrete reference class over patients with complete data, to which we can map patients with incomplete covariates at test time, and subsequently estimate treatment effects on the basis of these groups. 

Our contributions may thus be summarised as follows: We learn a discrete, low-dimensional, interpretable latent space representation of confounding. This representation allows us to learn equivalence classes among patients such that the specific causal effect of a patient can be approximated by the specific causal effect of the subgroups. We subsequently transfer this information to a set of test cases with incomplete measurements at test time such that we can estimate the causal effect. Finally, we demonstrate that our method outperforms existing approaches on established causal inference benchmarks and the real world applications for treating sepsis.

%% file: foundations.tex
\section{Preliminaries and Related Work}
\label{sec:related_work}

\paragraph{Potential Outcomes and Counterfactual Reasoning} Counterfactual reasoning (CR) has drawn large attention, particularly in the medical community. Such models are formalised in terms of \emph{potential outcomes} \cite{Neyman23,splawa1990application,rubin1978bayesian}. Assume we have two choices of taking a treatment $t$, and not taking a treatment (control) $c$. Let $Y_t$ denote the outcomes under $t$ and $Y_c$ denote outcomes under the control $c$. The counterfactual approach assumes that there is a pre-existing joint distribution $P(Y_t, Y_c)$ over outcomes. This joint distribution is hidden since $t$ and $c$ cannot be applied simultaneously.  Applying an action $t$ thus only reveals $Y_t$, but not $Y_c$. In this setting, computing the effect of an intervention involves computing the difference between when an intervention is made and when no treatment is applied \cite{pearl2009causality, morgan2015counterfactuals}. We would subsequently choose to treat with $t$ if, 
\begin{align}
\mathbb{E}[L(Y_t)] \leq \mathbb{E}[L(Y_c)] 
\end{align}
for loss $L$ over $Y_t$ and $Y_c$ respectively. Potential outcomes are typically applied to cross-sectional data \cite{NIPS2017_6767,DBLP:conf/nips/2017/SchulamS17} and sequential time settings. Notable examples of models for counterfactual reasoning include \cite{JohanssonTAR} and \cite{bottou2013counterfactual}. Specifically, \cite{JohanssonTAR} propose a neural network architecture called TARnet to estimate the effects of interventions. Similarly, Gaussian Process CR (GPCR) models are proposed in \cite{NIPS2017_6767, DBLP:conf/nips/2017/SchulamS17} and further extended to the multitask setting in \cite{AlaaS17}. Approaches that address missing data within the potential outcomes framework include \cite{cham2016propensity} and \cite{kallus2018causal}. The former adapts propensity score estimation for this purpose; the latter use low-rank matrix factorisation to deduce a set of confounders and compute treatment effects. Unlike all of these methods, we use of the IB criterion to learn treatment effects and adjust for confounding.
\paragraph{Decision-Theoretic View of Causal Inference} 
The decision theoretic approach to causal inference focuses on studying the \emph{effects of causes} rather than the causes of effects \cite{dawid07}. Here, the key question is \emph{what is the effect of the causal action on the outcome?} The outcome may be modelled as a random variable $Y$ for which we can set up a decision problem. That is, at each point, the value of $Y$ is dependent on whether $t$ or $c$ is selected. The decision-theoretic view of causal inference considers the distributions of outcomes given the treatment or control, $P_{t}$ and $P_{c}$ and explicitly computes an expected loss of $Y$ with respect to each action choice. Finally, the choice to treat with $t$ is made using Bayesian decision theory if,
\begin{align}
\mathbb{E}_{Y \sim P_t}[L(Y)] \leq \mathbb{E}_{Y \sim P_c}[L(Y)].
\end{align}
Thus in this setting, causal inference involves comparing the expected losses over the hypothetical distributions $P_t$ and $P_c$ for outcome $Y$.

\paragraph{Information Bottleneck}
The classical IB method \cite{art:tishby:ib} describes an information theoretic approach to compressing a random variable $X$ with respect to a second random variable $Y$. The compression of $X$ may be described by another random variable $Z$. Achieving an optimal compression requires solving the problem,
\begin{align}
{\min}_{p(z|x)} I(X;Z) - \lambda I(Z;Y),
\label{eq:1}
\end{align}
under the assumption that $Y$ and $Z$ are conditionally independent given $X$. That is, the classical IB method assumes that variables satisfy the Markov relation $Z - X - Y$. $I$ in Eqn. \ref{eq:1} represents the mutual information between two random variables and $\lambda$ controls the degree of compression. In its classical form, the IB principle is defined only for discrete random variables. However in recent years multiple IB relaxations and extensions, such as for Gaussian  \cite{art:chechik:gib} and meta-Gaussian variables \cite{mgib}, have been proposed. Among these extensions, is the latent variable formulation of the IB method (e.g. \cite{DVIB, Wieczorek}). Here, one assumes structural equations of the form,
\begin{align}
z = f(x) + \eta_z, \\
y = g(z) + \eta_y. 
\end{align}
These equations give rise to a different conditional independence relation,  $X - Z - Y$. While both independences cannot hold in the same graph, in the limiting case where the noise term $\eta_y \rightarrow 0$, $Z \bigCI X \mid Y$. In what follows, we assume the latent variable formulation of the IB.

\paragraph{Deep Latent Variable Models} 
Deep latent variable models have recently received remarkable attention and been applied to a variety of problems. Among these, variational autoencoders (VAEs) employ the reparameterisation trick introduced in \cite{kingma2013auto, pmlr-v32-rezende14} to infer a variational approximation over the posterior distribution of the latent space $p(z|x)$.   Important work in this direction include \cite{KingmaSemi} and \cite{Jang}. Most closely related to the work we present here, is the application of VAEs in a healthcare setting by \cite{Louizos}. Here, the authors introduce a Cause-Effect VAE (CEVAE) to estimate the causal effect of an intervention in the presence of noisy proxies. Despite their differences, it has been shown that there are several close connections between the VAE framework and the previously described latent variable formulation of the IB principle. This is essentially a VAE where $X$ is replaced by $Y$ in the decoder with $\lambda = 1$. In contrast, the approach in this paper considers the IB principle to perform causal inference in scenarios where our data has a systematic missingness at test time.

%% file: method.tex
\section{Method}
\label{sec:method}
\begin{figure}[!h]
    \centering
        \begin{subfigure}{0.25\textwidth}
        		\centering
        		\includegraphics[width=\textwidth]{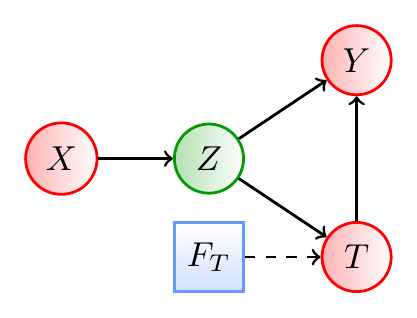}
        		\caption{}
        		\label{fig:cases}
        \end{subfigure}         
        \hspace{0.5cm}
        \begin{subfigure}{0.35\textwidth}
                \centering
        		\includegraphics[width=\textwidth]{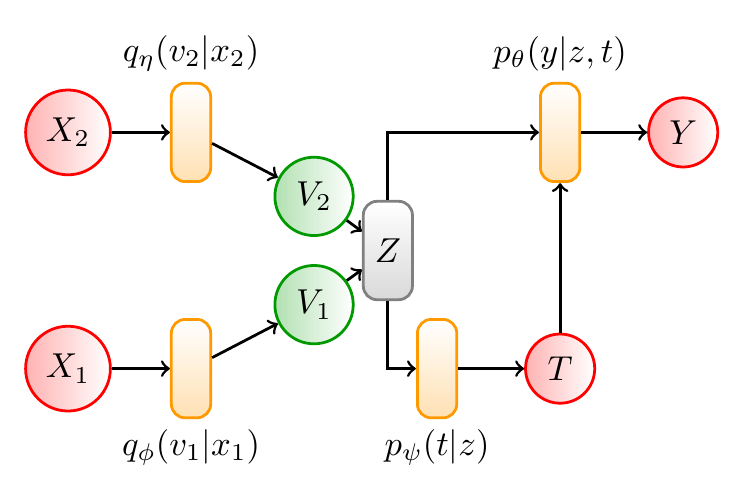}
        		\caption{}
        		\label{model-img}
        \end{subfigure}  
    \caption{(a) Influence diagram of the CEIB. Red and green circles correspond to observed and latent random variables respectively, while blue rectangles represent interventions. We identify a low-dimensional representation $Z$ of covariates $X$ to estimate the effects of an intervention on outcome $Y$ where incomplete covariate information is available at test time. (b) Graphical illustration of the CEIB. Orange rectangles represent deep networks parameterising the random variables.}  
\end{figure}

In this section, we present an approach based on the IB principle for estimating the causal effects of an intervention with incomplete covariate information. We refer to this model as a \emph{Cause-Effect Information Bottleneck (CEIB)}. In recent years, there has been a growing interest in the connections between the IB principle and deep neural networks \cite{TishbyZ15,  DVIB, Wieczorek}. Here, we use the non-linear expressiveness of neural networks with the IB criterion to learn a sufficiently reduced representation of confounding, based on which we can approximate the effects of an intervention more effectively. Specifically, we interpret our model from the decision-theoretic view \cite{dawid07} of causal inference.  
\paragraph{Problem Formulation} Like other approaches in the decision-theoretic setting, our goal is to estimate the \textit{Average Causal Effect (ACE)} of $T$ on $Y$. If we assume that $F_T=0$ or $F_T=1$ define the interventional regimes, and $F_T=\emptyset$ the observational regime, the ACE is given by,
\begin{align} 
\label{eq1}
	ACE :&= \mathbb{E}[Y \mid F_T=1] - \mathbb{E}[Y \mid F_T=0].
\end{align}
The ACE in Equation \ref{eq1} is defined in terms of the interventional regime however, in practice we can only collect data on $F_T = \emptyset$. The observational counterpart of the ACE formally be defined as:
\begin{align}
\label{eq2}
	ACE :&= \mathbb{E}[Y \mid T=1, F_T=\emptyset] - \mathbb{E}[Y \mid T=0, F_T=\emptyset].
\end{align}
In general, the ACE in Equations \ref{eq1} and \ref{eq2} are not equal unless we assume ignorable treatment assignments or under the conditional independence assumption $Y \indep F_T \mid T$. This assumption expresses that the distribution of $Y \mid T$ is the same in the interventional and observational regimes. In the presence of confounding, the treatment assignment $F_T$ may only be ignored when estimating $Y$, provided a sufficient covariate $Z$ and $T$ \cite{dawid07}. That is, $Z$ is a sufficient covariate for the effect of $T$ on outcome $Y$ if $Z \indep  F_T$ and $Y \indep F_T \mid (Z,T)$. In this case, it can be shown by Pearl's backdoor criterion  \cite{pearl2009causality} that the ACE may be defined in terms of the Specific Causal Effect (SCE),
\begin{align} \label{eq5}
\begin{split}
   ACE :&= \mathbb{E}[SCE(Z) \mid F_T=\emptyset]
\end{split}
\end{align}
where
\begin{align} \label{eq6}
\begin{split}
   SCE(Z) :&= \mathbb{E}[Y \mid Z, T=1, F_T=\emptyset] - \mathbb{E}[Y \mid Z, T=0, F_T=\emptyset].
\end{split}
\end{align}
We employ the following assumptions and notation. Let $X =  (X_1, X_2)$ denote a set of patient covariates based on which we would like to estimate treatment effects. During training, we assume that all covariates $X \in \mathbb{R}^d$ can be observed as in a medical study, where dimension $d$ is large. Outside the study at test time however, we assume covariates $X_1$ are not usually observed, e.g. due to the expensive data acquisition process. That is, we assume the same feature dimensions are missing for all 
patients at testing. Let $Y \in \mathbb{R}$ denote the outcomes following treatments $T$. For simplicity and ease of comparison with prior methods on existing benchmarks, we consider treatments $T$ that are binary, but our method is applicable for any general $T$. We assume strong ignorability or that all confounders are measured for a set of patients. The causal graph of our model is shown in Figure \ref{fig:cases}. Importantly, estimating the ACE in this case only requires computing a distribution $Y|Z, T$, provided $Z$ is a sufficient covariate. In what follows, we use the IB to learn a sufficient covariate that allows us to approximate the distribution $Y \mid Z, T$ in Figure \ref{fig:cases}. 

\paragraph{Performing a Sufficient Reduction of the Covariate} We propose modelling this task with an extended formulation of the information bottleneck using the architecture proposed in Figure \ref{model-img}. The IB approach allows us to learn a low-dimensional interpretable compression of the relevant information during training, which we can use to infer treatment effects where covariate information is incomplete at test time. 

We adapt the IB for learning the outcome of a therapy when incomplete covariate information is available for $X_2$ at test time. To do so, we consider the following extended parametric form of the IB, 
\begin{align}
\max_{\phi, \theta, \psi, \eta} -I_{\phi} (V_1; X_1) - I_{\eta}(V_2; X_2) + \lambda I_{\phi, \theta, \psi, \eta}(Z; (Y, T)), 
\label{criterion}
\end{align}
where $V_1$ and $V_2$ are low-dimensional discrete representations of the covariate data, $Z = (V_1, V_2)$ is a concatenation of $V_1$ and $V_2$ and $I$ represents the mutual information parameterised by networks $\phi$, $\psi$, $\theta$, and $\eta$ respectively. We assume a parametric form of the conditionals $q_\phi(v_1|x)$, $q_\eta(v_2|x)$, $p_\theta(y|t, z)$, $p_\psi(t|z)$. The first two terms in Equation \ref{criterion} for our encoder model have the following forms:
\begin{align}
I_\phi(V_1; X_1) &= D_{KL}(q_\phi(v_1|x_1)p(x_1)||p(v_1)p(x_1)) = \mathbb{E}_{p(x_1)}D_{KL}(q_\phi(v_1|x_1)||p(v_1)) \\
I_\eta(V_2; X_2) &= D_{KL}(q_\eta(v_2|x_2)p(x_2)||p(v_2)p(x_2)) = \mathbb{E}_{p(x_2)}D_{KL}(q_\eta(v_2|x_2)||p(v_2))
\end{align}
The decoder model in Equation \ref{criterion} can analogously be expressed as:
\begin{align}
I_{\phi, \theta, \psi, \eta}(Z; (Y, T)) 
&\geq \mathbb{E}_{p(x, y, t)} \mathbb{E}_{p_{\phi,\eta}(z|x)} \big[\log p_\theta(y|t, z) + \log p_\psi(t|z)\big] + h(y,t)\text{\footnotemark}.
\end{align}
\footnotetext{The lower bound follows from the fact that the mutual information between $Z$ and $Y, T$ can be expressed as a sum of the expected value of $\log p_\theta(y|t, z) + \log p_\psi(t|z)$, entropy $h(y,t)$ and two KL-divergences, which are by definition non-negative.}where $h(y,t) = -\mathbb{E}_{p(y,t)} \log p(y,t)$ is the entropy of $(y,t)$. For the decoder model, we use an architecture similar to the TARnet \cite{JohanssonTAR}, where we replace conditioning on high-dimensional covariates $X$ with conditioning on reduced covariate $Z$. We can thus formulate the conditionals as, 
\begin{align}
p_\psi(t|z) = \textup{Bern}(\sigma(f_1(z)))\hspace{0.5cm} \nonumber\\
p_\theta(y|t, z) = \mathcal{N}(\mu = \hat\mu, \varsigma^2 = \hat{s}),
\end{align}
with logistic function $\sigma(\cdot)$, and outcome $Y$ given by a Gaussian distribution parameterised with a TARnet with $\hat\mu = t f_2(z) + (1- t)f_3(z)$. Note that the terms $f_k$ correspond to neural networks. While distribution $p(t|z)$ is included to ensure the joint distribution over treatments, outcomes and covariates is identifiable, in practice, our goal is to approximate the effects of a given $T$ on $Y$. Hence, we train our model in a teacher forcing fashion by using the true treatment assignments $T$ from the data and fixing the $T$s at test time. Unlike other approaches for inferring treatment effects, the Lagrange parameter $\lambda$ in the IB formulation in Equation \ref{criterion} allows us to adjust the degree of compression, which, in this context, enables us to learn a sufficient statistic $Z$. In particular, adjusting $\lambda$ enables us to explore a range of such representations from having a completely insufficient covariate to a completely sufficient compression of confounding.

\paragraph{Learning Equivalence Classes and Distribution Transfer} Since $V_1$ and $V_2$ are discrete latent representations of the covariate information, we make use of the Gumbel softmax reparameterisation trick \cite{Jang} to draw samples $Z$ from a categorical distribution with probabilities $\pi$. Here,
\begin{align}
z = \texttt{one\_hot} (\arg \max_i [g_i + \log \pi_i]),
\label{Gumbel-max}
\end{align}
where $g_1, g_2, \ldots, g_k$ are samples drawn from Gumbel(0,1). The softmax function is used to approximate the $\arg \max$ in Equation \ref{Gumbel-max}, and generate $k$-dimensional sample vectors $w \in \Delta^{k-1}$, where
\begin{align}
w_i = \frac{\exp((\log(\pi_i) + g_i)/\tau)}{\sum_{j=1}^{k} \exp((\log(\pi_j) + g_j)/\tau)},  i = 1, \ldots, k.
\end{align}
and $\tau$ is the softmax temperature parameter. By using the Gumbel softmax reparameterisation trick to obtain a discrete representation of relevant information, we can learn equivalence classes among patients based on which we can compute the SCE for each group using sufficient covariate $Z$ via Equation \ref{eq6}. Specifically, during training, $X_1$ and $X_2$ are used to learn cluster assignment probabilities $\pi$ for each data point. At test time, we subsequently assign an example with missing covariates to the relevant equivalence class. Computing the SCE allows us potentially to tailor treatments to specific groups based on $Z$ rather than an entire population --- an important aspect in healthcare where patients are typically heterogeneous. Based on the SCE, we can also compute the population-level effects of an intervention via the ACE from Equation \ref{eq5}. In the absence of the latent compression via CEIB and the discrete representation of relevant information, it would not be possible to transfer knowledge from examples with complete information to cases with incomplete information, since estimating treatment effects would require integrating over all covariates --- an infeasible task in high dimensions.


%% file: results.tex
\section{Experiments}
\label{sec:experiments}
The goal of the experiments is to demonstrate the ability of CEIB to accurately infer treatment effects while learning a low-dimensional interpretable representation of confounding in cases where partial covariate information is available at test time. We report the ACE and SCE values in our experiments for this purpose. In general, the lack of ground truth in real world data makes evaluating causal inference algorithms a difficult problem. To overcome this, in our artificial experiments we consider a semi-synthetic data set where the true outcomes and treatment assignments known\footnote{Additional experiments can be found in the supplement}.

\paragraph{The Infant Health and Development Program:} The Infant Health and Development Program (IHDP) \cite{IHDP, JHill} is a randomised control experiment assessing the impact of educational intervention on outcomes of pre-mature, low birth weight infants born in 1984-1985. Measurements from children and their mothers were collected to study the effects of childcare and home visits from a trained specialist on test scores. The data includes treatment groups, health indices, mothers' ethnicity and educational achievement. \cite{JHill} extract features and treatment assignments from the real-world clinical trial, and introduce selection bias to the data artificially by removing a non-random portion of the treatment group, in particular children with non-white mothers. In total, the data set consists of 747 subjects (139 treated, 608 control), each represented by 25 covariates measuring characteristics of the child and their mother. The data set is divided into 60/10/30\% into training/validation/testing sets. For our setup, we use encoder and decoder architectures with 3 hidden layers. Our model is trained with Adam optimiser \cite{KingmaB14} with a learning rate of 0.001. We compare the performance of CEIB for predicting the ACE against several existing baselines as in \cite{Louizos}\footnote{OLS-1 is a least squares regression; OLS-2 uses two separate least squares regressions to fit the treatment and control groups respectively; TARnet is a feedforward neural network from \cite{pmlr-v70-shalit17a}; KNN is a $k$-nearest neighbours regression; RF is a random forest; BNN is a balancing neural network \cite{JohanssonTAR}; BLR is a balancing linear regression \cite{JohanssonTAR}, and CFRW is a counterfactual regression that using the Wasserstein distance \cite{pmlr-v70-shalit17a}.}. We train our model with $k = 4$, $d = 3$-dimensional Gaussian mixture components. 

\begin{table}[!h]
\begin{subtable}{0.45\textwidth}
  \centering
  \begin{tabular}{lclc|c|}
    \hline
    Method  & $\epsilon_{ACE}^{within-s}$  & $\epsilon_{ACE}^{out-of-s}$\\
    \hline
    OLS-1  & $.73\pm.04$  & $.94\pm.06$ \\
    OLS-2 &  $.14\pm.01$  & $.31\pm.02$ \\
    KNN & $.14\pm.01$  & $.79\pm.05$  \\
    BLR &  $.72\pm.04$  & $.93\pm.05$\\
    TARnet & $.26\pm.01$ & $.28\pm.01$ \\
    BNN  & $.37\pm.03$ & $.42\pm.03$ \\
    RF  & $.73\pm.05$  & $.96\pm.06$ \\
    CEVAE  & $.34\pm.01$  & $.46\pm.02$ \\
    CFRW  & $.25\pm.01$ & $.27\pm.01$\\
    \hline
    CEIB  & $\mathbf{.11\pm.01}$  & $\mathbf{.21\pm.01}$ \\
    \hline
  \end{tabular}
  \caption{}
\end{subtable}
\begin{subtable}{0.45\textwidth}
  \centering
  \begin{tabular}{lcc}
    \hline
    Method  & $\epsilon_{ACE}^{within-s}$  & $\epsilon_{ACE}^{out-of-s}$\\
    \hline
    TARnet  & $.30\pm.01$  & $.34\pm.01$ \\
    CFRW  & $.28\pm.01$ & $.49\pm.02$\\
    \hline
    CEIB  & $\mathbf{.14\pm.02}$  & $\mathbf{.23\pm.01}$ \\
    \hline
  \end{tabular}
  \caption{}
\end{subtable}
\caption{\label{table:ihdp}(a) Within-sample and out-of-sample mean and standard errors in ACE across models on the complete IHDP data set. A smaller value indicates better performance. Bold values indicate the method with the best performance. (b) Within-sample and out-of-sample mean and standard errors in ACE across models using a reduced set of 22 covariates at test time.}
\end{table}

\paragraph{Experiment 1:} In the first experiment, we compared the performance of CEIB for estimating the ACE against the baselines when using the complete set of measurements at test time. These results are shown in Table \ref{table:ihdp}a. Evidently, CEIB outperforms existing approaches. To demonstrate that we can transfer the relevant information to cases where covariates are incomplete at test time, we artificially excluded $n = 3$ covariates that have a moderate correlation with ethnicity at test time. We compute the ACE and compare this to the performance of TARnet and CFRW also on the reduced set of covariates (Table\ref{table:ihdp}b). If we extend this to the extreme case of removing 8 covariates at test time, the out-of-sample error in predicting the ACE increases to 0.29 +/- 0.02. Thus CEIB achieves state-of-the-art predictive performance for both in-sample and out-of-sample predictions, even with incomplete covariate information.  

\begin{figure}[!h]
    \centering
        \begin{subfigure}{0.3\textwidth}
        		\centering
        		\includegraphics[width=\textwidth]{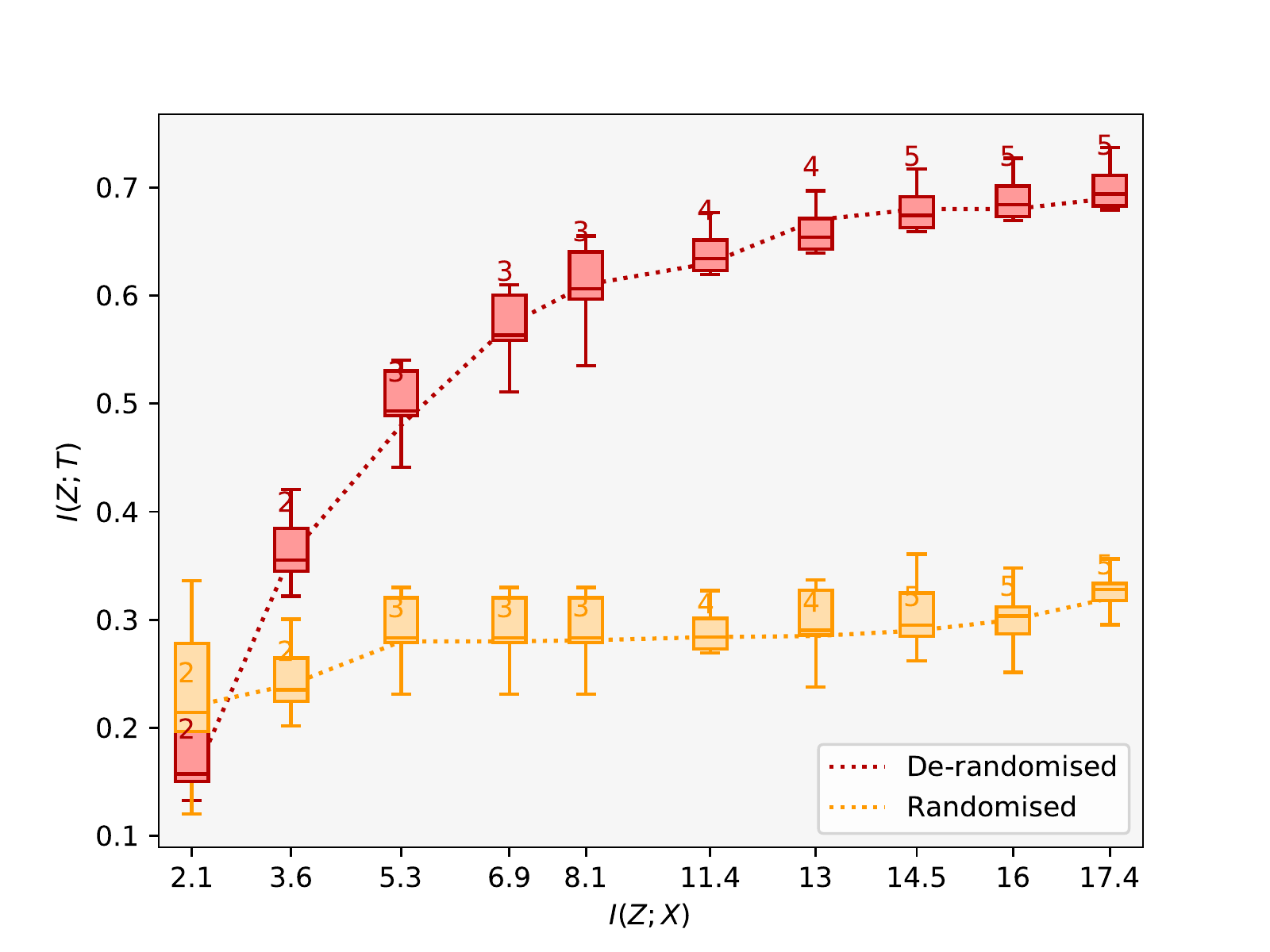}
        		\caption{}
        		\label{information curve:b}
        \end{subfigure}         
        \begin{subfigure}{0.3\textwidth}
                \centering
        		\includegraphics[width=\textwidth]{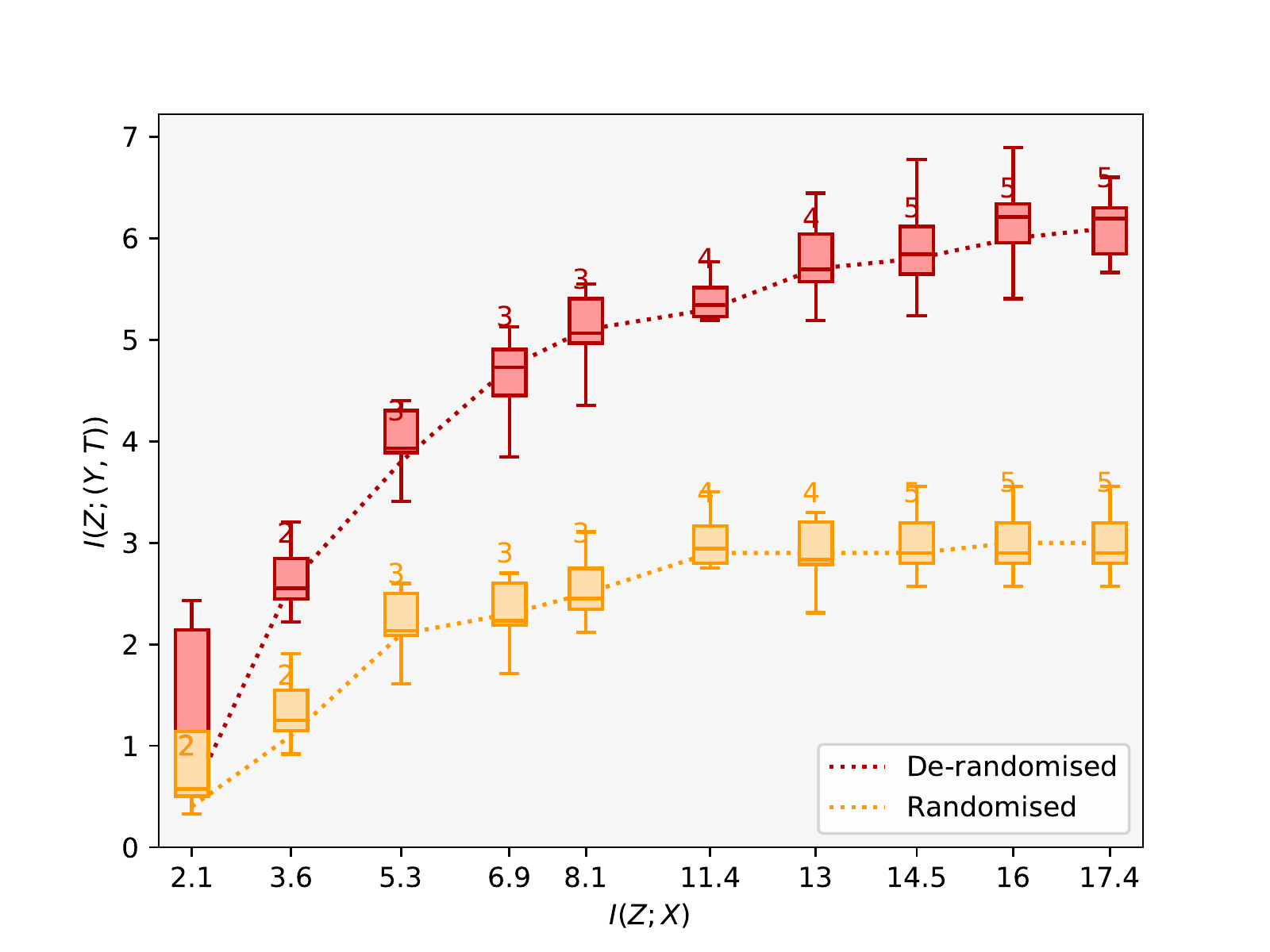}
        		\caption{}
        		\label{information curve:a}
        \end{subfigure}  
        \begin{subfigure}{0.3\textwidth}
        \centering
        \includegraphics[width=\textwidth]{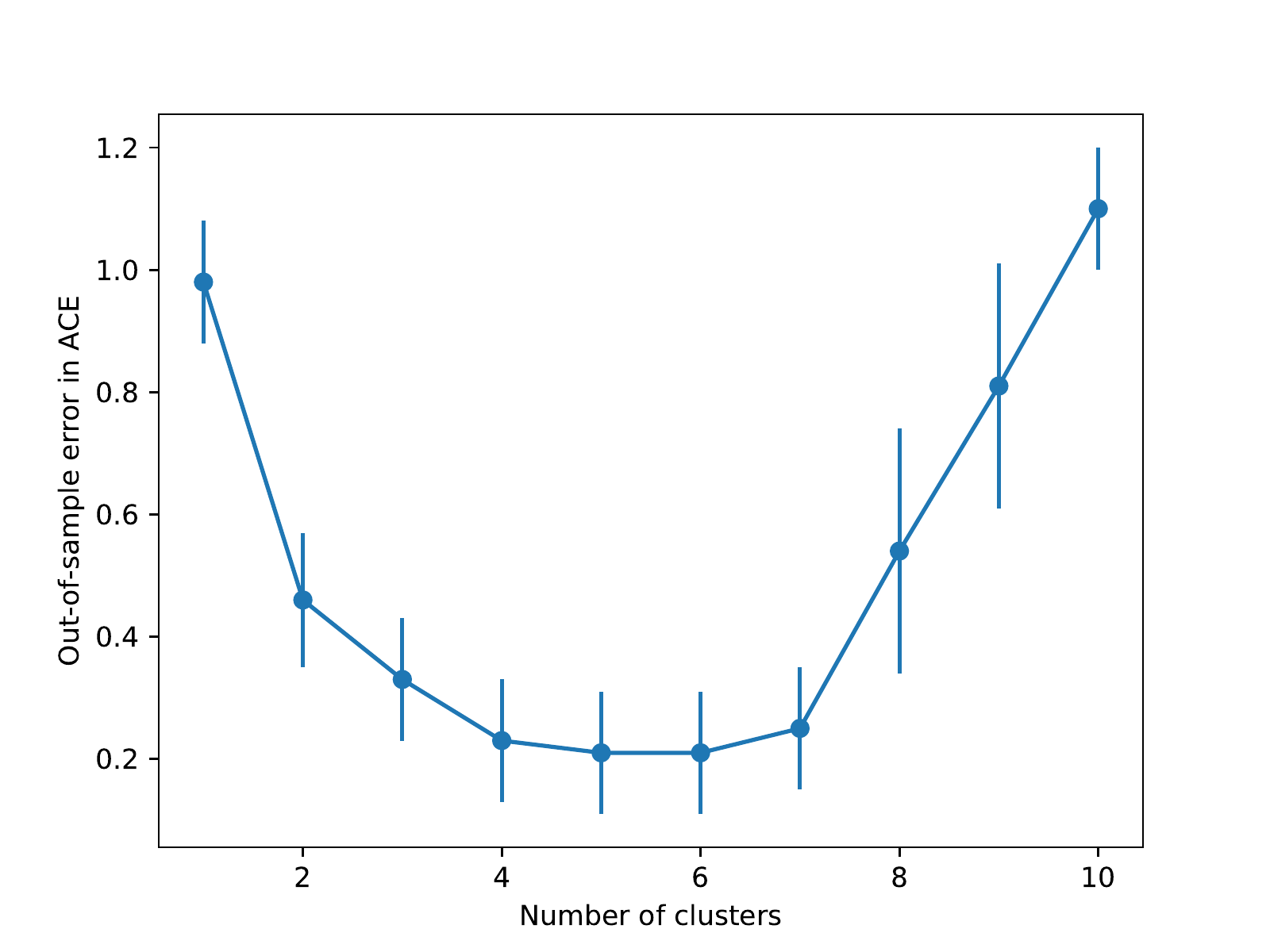}
        \caption{}
        \label{fig:ATE-error-cluster}
        \end{subfigure}
    \caption{(a) Information curves for $I(Z; T)$ and (b) $I(Z;(Y, T))$ with de-randomised and randomised data respectively. When the data is randomised, the value of $I(Z;T)$ is  close to zero. The differences between the curves illustrates confounding. When data is de-randomised, we are able to estimate treatment effects more accurately by accounting for this confounding. (c) Out-of-sample error in ACE with a varying number of clusters.}  
\end{figure}

\paragraph{Experiment 2:} Building on Experiment 1, we perform an analysis of the latent space of our model to assess whether we learn a sufficiently reduced covariate. We use the IHDP data set as before, but this time consider both the data before introducing selection bias (analogous to a randomised study), as well as after introducing selection bias by removing a non-random proportion of the treatment group as before (akin to a de-randomised study). We plot the information curves illustrating the number of latent dimensions required to reconstruct the output for the terms $I(Z; (Y, T))$ and $I(Z, T)$ respectively for varying values of $\lambda$. These results are shown in Figure \ref{information curve:b} and \ref{information curve:a}. Theoretically, we should be able to examine the shape of the curves to identify whether a sufficiently reduced covariate has been obtained. In particular, where a study is randomised, the sufficient covariate $Z$ should have no impact on the treatment $T$. In this case, the mutual information $I(Z, T)$ should be approximately zero and the curve should remain flat for varying values of $I(Z, X)$. This result is confirmed in Figure \ref{information curve:b}. The information curves in Figure \ref{information curve:a} additionally demonstrate our model's ability to account for confounding when predicting the overall outcomes. Specifically, the point at which each of the information curves saturates is the point at which we have learnt a sufficiently reduced covariate based on which we can infer treatment effects. The curves also highlight benefit of adjusting $\lambda$, since we obtain a task-dependent adjustment of the latent space which allows us to explore a range of solutions, from having a completely insufficient covariate to a completely sufficient compression of the covariates where the information curve saturates. Overall, we are able to learn a low-dimensional representation that is consistent with the ethnicity confounder and account for its effects when predicting treatment outcomes.

\begin{figure}[!h]
    \centering
        \begin{subfigure}{0.20\textwidth}
        		\centering
        		\includegraphics[width=\textwidth]{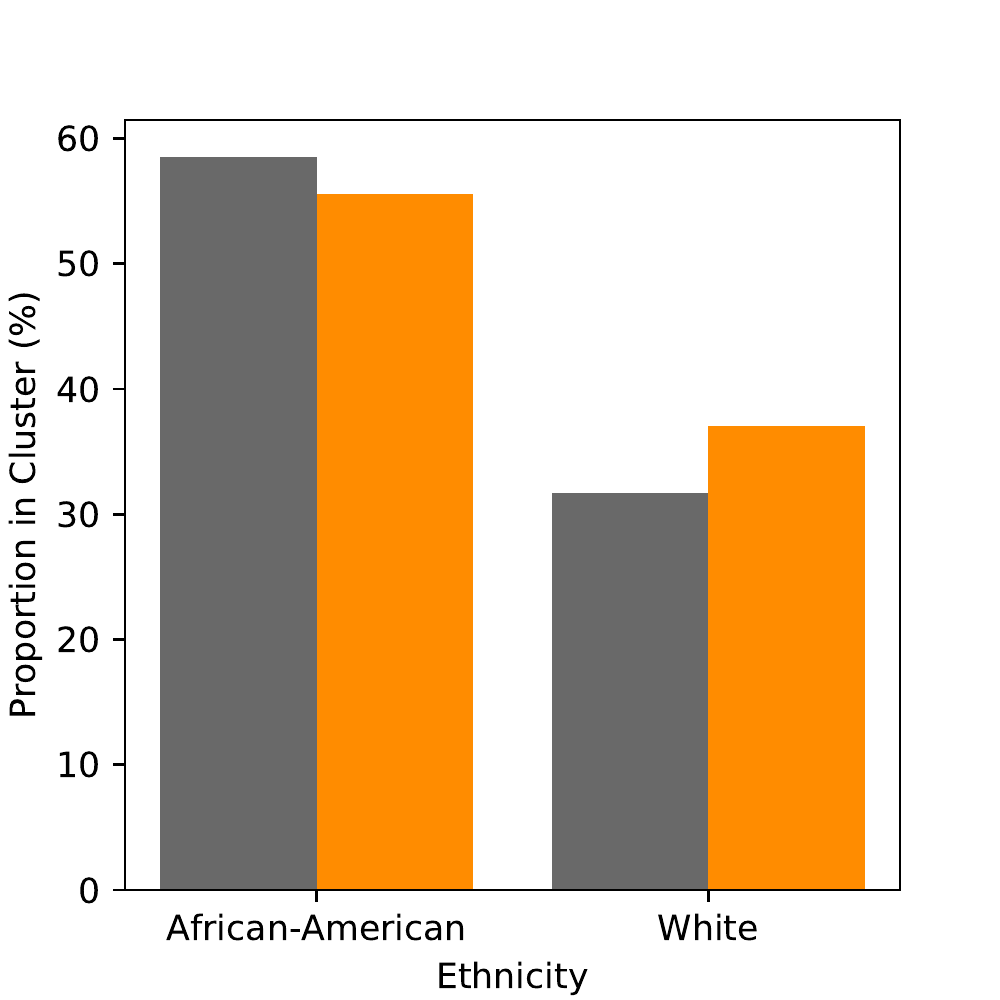}
        		\caption{SCE: 4.9 }
        		\label{cluster1}
        \end{subfigure}         
        \begin{subfigure}{0.20\textwidth}
        		\centering
        		\includegraphics[width=\textwidth]{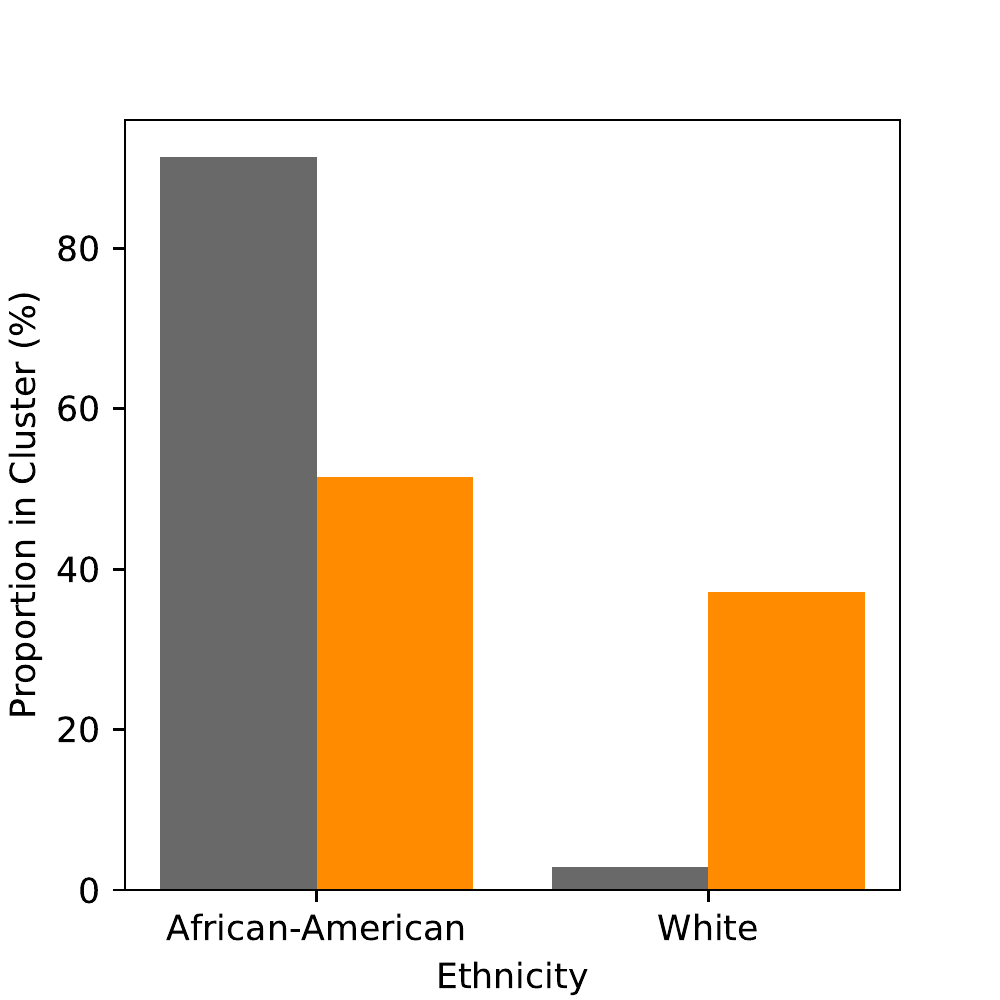}
        		\caption{SCE: 2.7}
        		\label{cluster2}
        \end{subfigure}
        \begin{subfigure}{0.20\textwidth}
                \centering
        		\includegraphics[width=\textwidth]{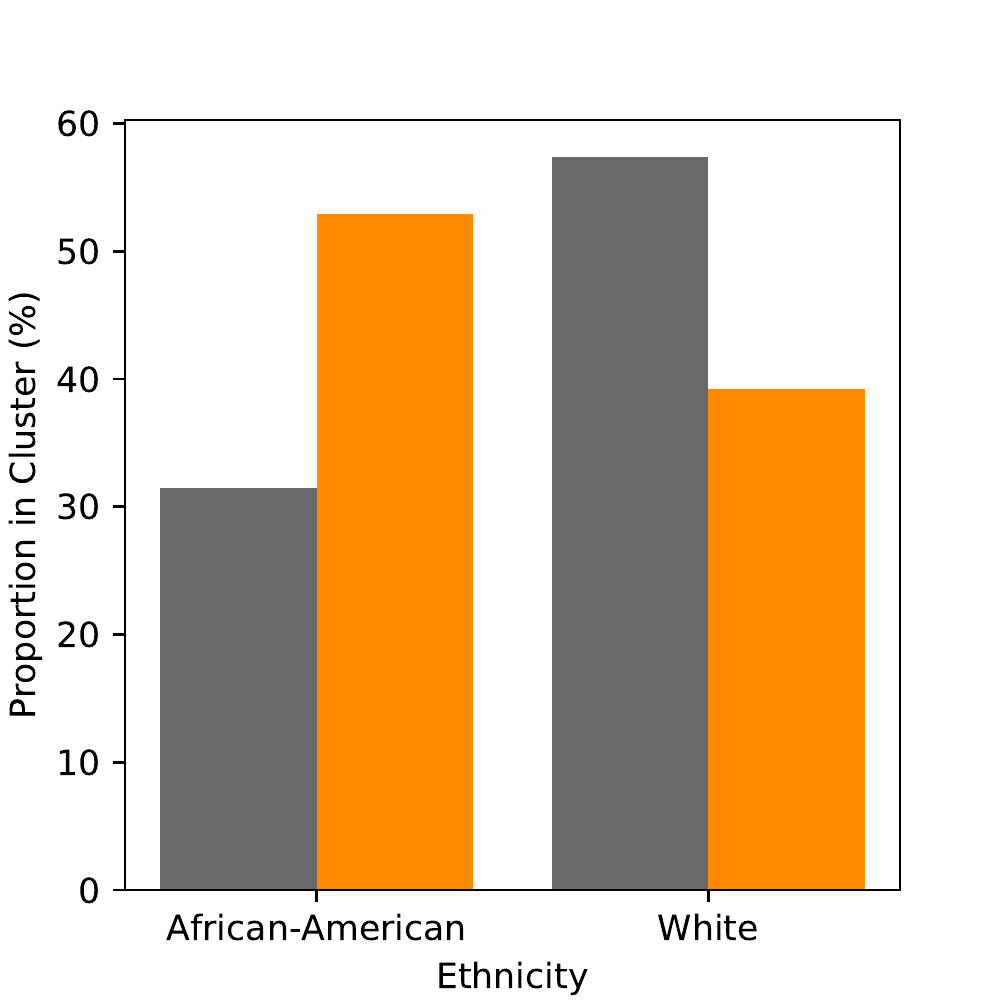}
        		\caption{SCE: 4.3}
        		\label{cluster3}
        \end{subfigure}  
             \begin{subfigure}{0.20\textwidth}
                \centering
        		\includegraphics[width=\textwidth]{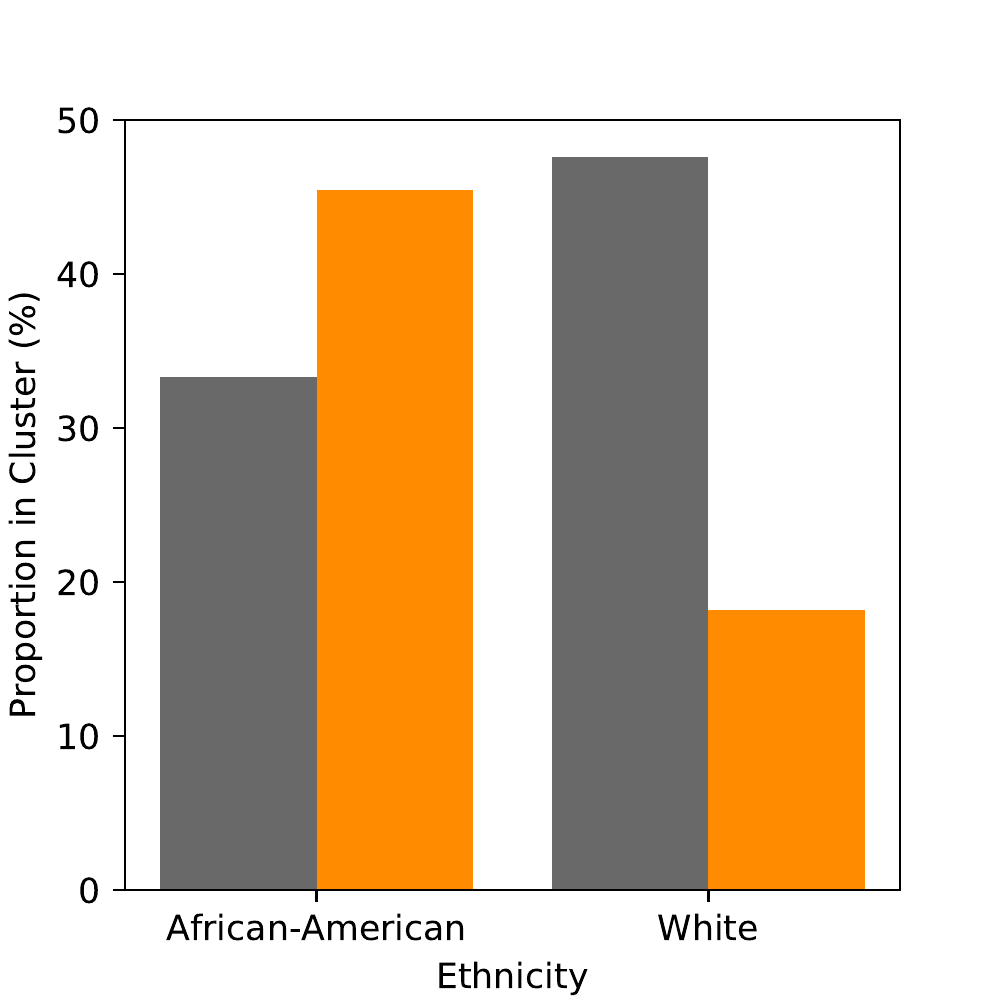}
        		\caption{SCE: 4.1}
        		\label{cluster4}
        \end{subfigure}  
    \caption{\label{fig:cluster} Illustration of the proportion of major ethnic groups within the four clusters. Grey and orange indicate de-randomised and randomised data respectively. The first cluster in (a) is a neutral cluster. The second cluster in (b) shows an enrichment of information in the African-American group. Clusters 3 and 4 in (c) and (d) respectively, show an enrichment of information in the White group.}
\end{figure}

We also analysed the discretised latent space by comparing the proportions of ethnic groups of test subjects in each cluster in the de-randomised setting. 
These results are shown in Figure \ref{fig:cluster} where we plot a hard assignment of test subjects to clusters on the basis of their ethnicity. Evidently, the clusters exhibit a clear structure with respect to ethnicity. In particular, Cluster 2 in Figure \ref{cluster2} has a significantly higher proportion of non-white members in the de-randomised setting. The discretisation also allows us to calculate the SCE for each cluster. In general, Cluster 2 tends to have a lower SCE than the other groups. This is consistent with how the data was de-randomised, since we removed a proportion of the treated instances with non-white mothers. Conditioning on this kind of information is thus crucial to be able to accurately assess the impact of educational intervention on test scores. Finally, we assess the error in estimating the ACE when varying the number of mixture components in Figure \ref{fig:ATE-error-cluster}.  When the number of clusters is larger, the clusters get smaller and it becomes more difficult to reliably estimate the ACE since we average over the cluster members to account for partial covariate information at test time. Here, model selection is made by observing where the error in estimating the ACE stabilises (anywhere between 4-7 mixture components).

\begin{figure}[!h]
    \centering
        \begin{subfigure}{0.3\textwidth}
        		\centering
        		\includegraphics[width=\textwidth]{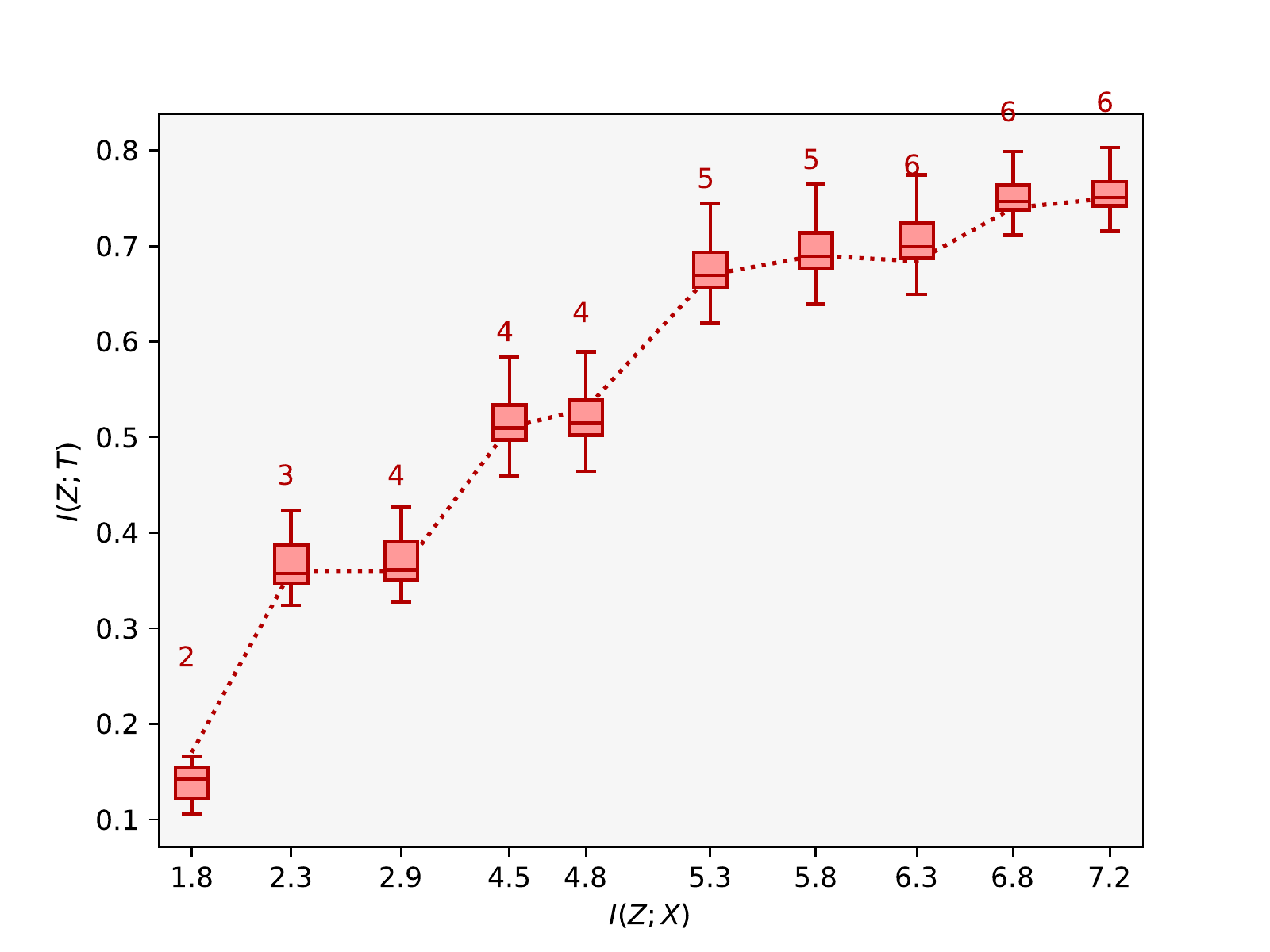}
        		\caption{}
        		\label{information curve:c}
        \end{subfigure}         
        \begin{subfigure}{0.3\textwidth}
                \centering
        		\includegraphics[width=\textwidth]{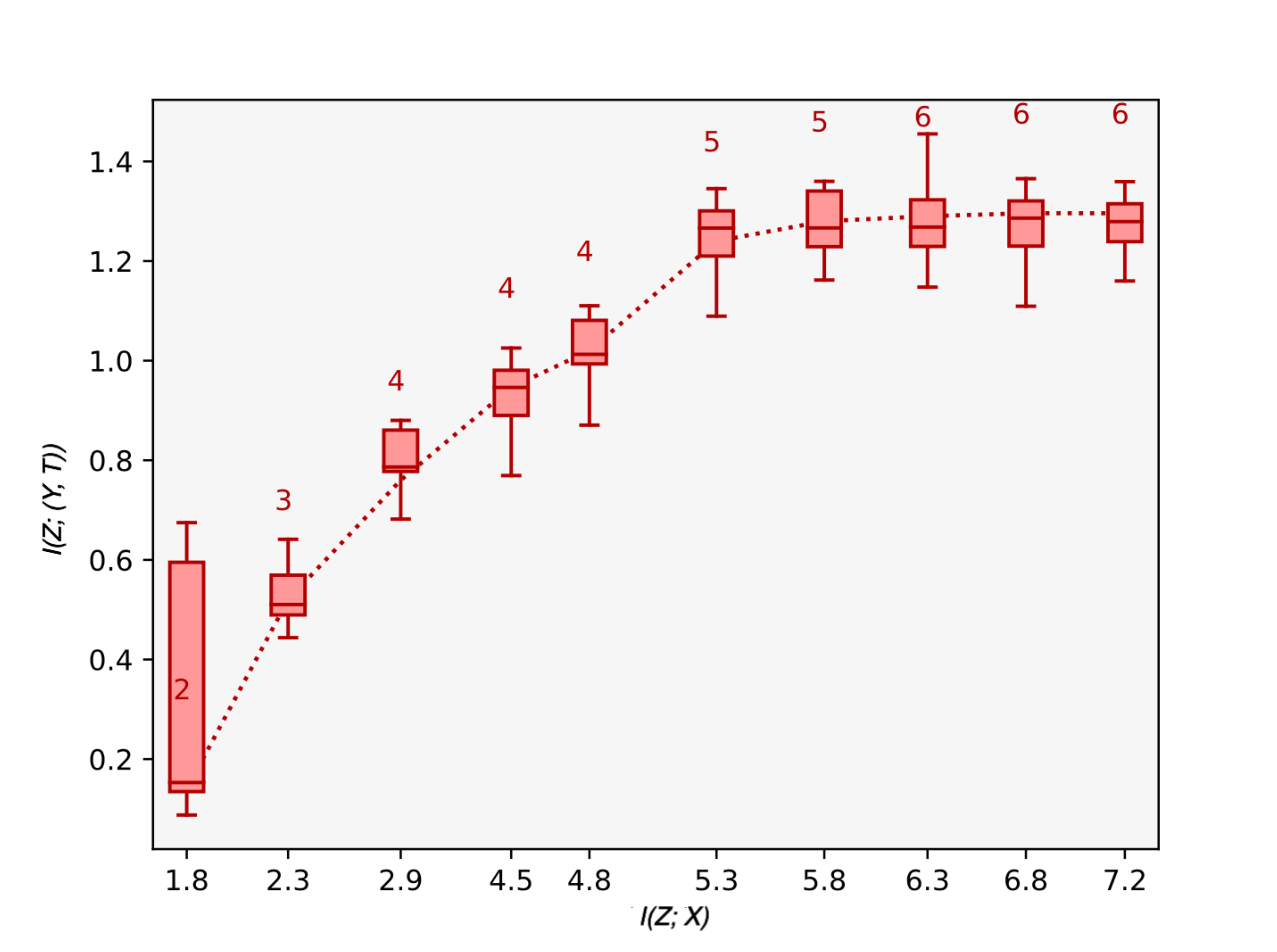}
        		\caption{}
        		\label{information curve:d}
        \end{subfigure}  
    \caption{Subfigures (a) and (b) illustrate the information curve $I(Z; T)$ and $I(Z; (Y; T))$ for the task of managing sepsis. We perform a sufficient reduction of the covariates to 6-dimensions and are able to approximate the ACE on the basis of this.}  
\end{figure}

\paragraph{Sepsis Management:} We illustrate the performance CEIB on the real-world task of managing and treating sepsis. Sepsis is one of the leading causes of mortality within hospitals and treating septic patients is highly challenging, since outcomes vary with interventions and there are no universal treatment guidelines. We use data from the Multiparameter Intelligent Monitoring in Intensive Care (MIMIC-III) database \cite{johnson2016mimic}. We focus  on patients satisfying Sepsis-3 criteria (16 804 patients in total). For each patient, we have a 48-dimensional set of physiological parameters including demographics, lab values, vital signs and input/output events.  Our outcomes $Y$ correspond to the odds of mortality, while we binarise medical interventions $T$ according to whether or not a vasopressor is administered.  The data set is divided into 60/20/20\% into training/validation/testing sets. We train our model with six 4-dimensional Gaussian mixture components and analyse the information curves and cluster compositions respectively. 
\begin{figure*}[!h]
    \centering
        \begin{subfigure}{0.15\textwidth}
        		\centering
        		\includegraphics[width=\textwidth]{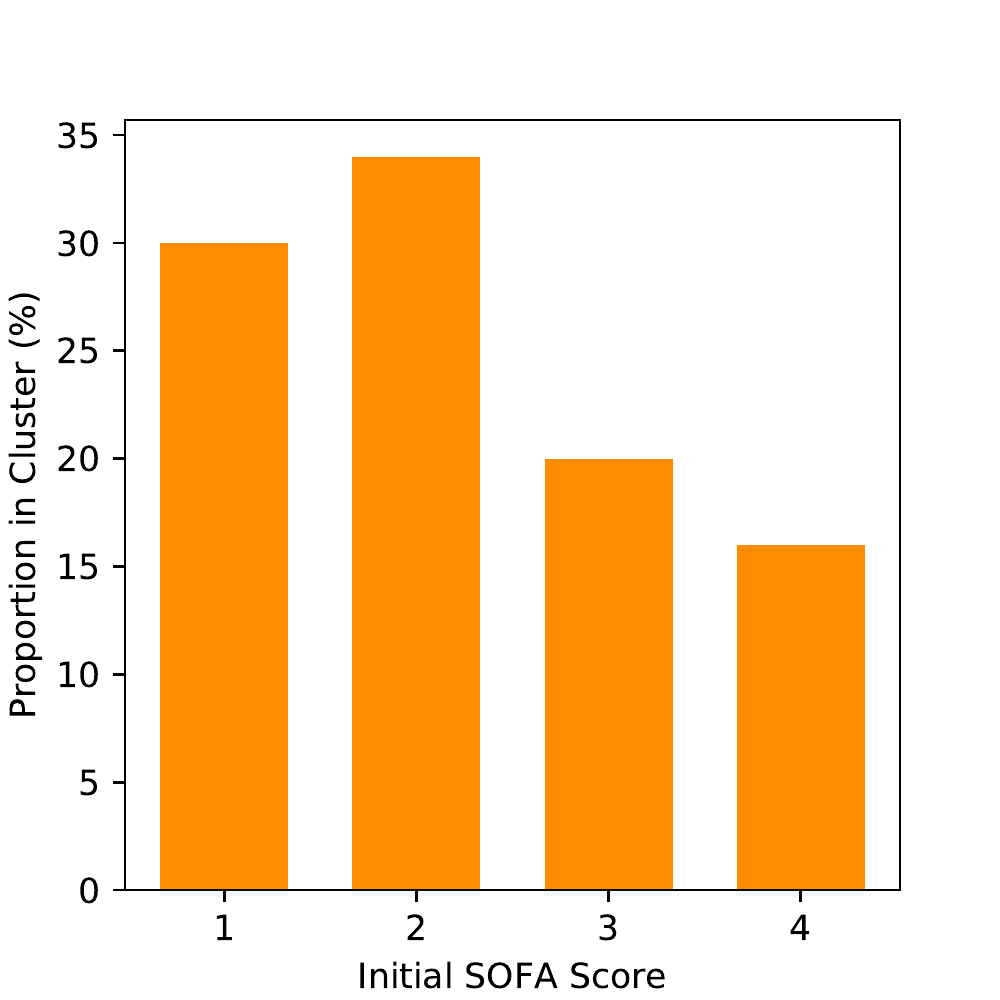}
        		\caption{SCE: -1.2}
        		\label{cluster1s}
        \end{subfigure}         
        \begin{subfigure}{0.15\textwidth}
        		\centering
        		\includegraphics[width=\textwidth]{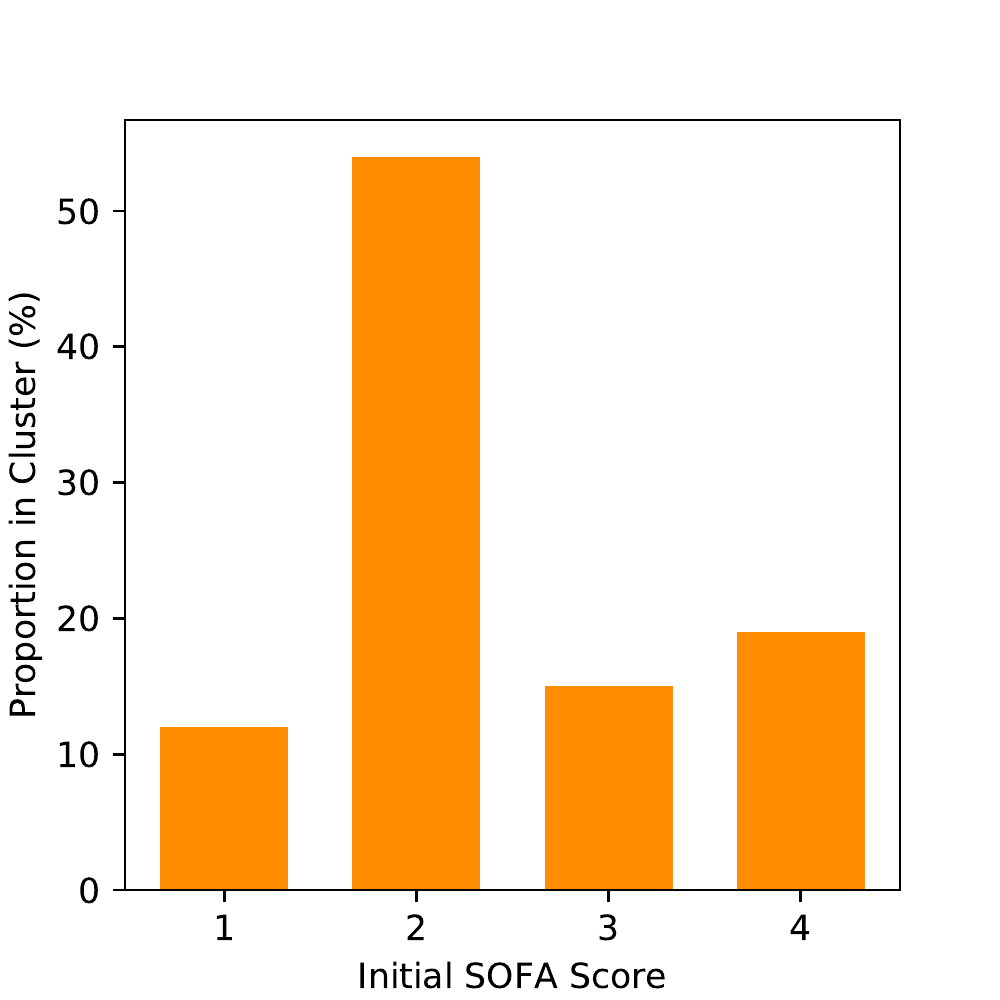}
        		\caption{SCE: -2.7 }
        		\label{cluster2s}
        \end{subfigure}
        \begin{subfigure}{0.15\textwidth}
                \centering
        		\includegraphics[width=\textwidth]{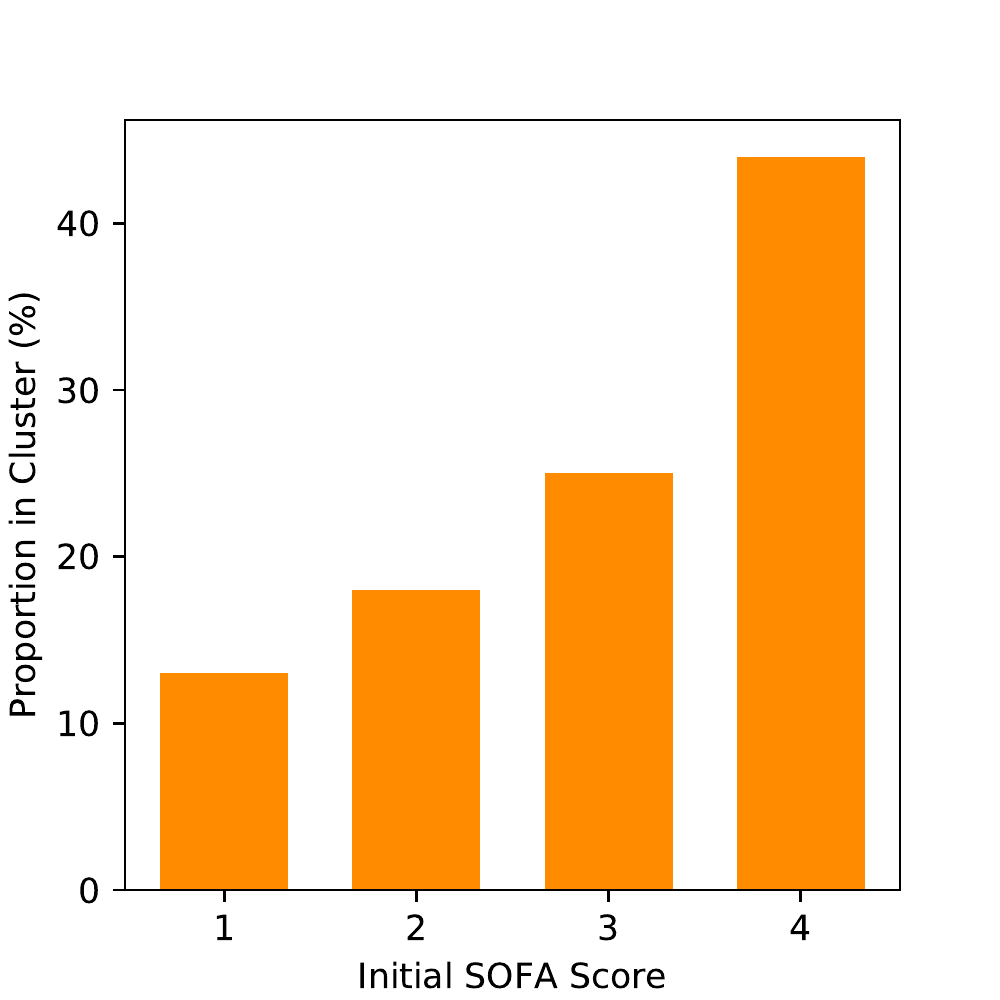}
        		\caption{SCE: 1.4 }
        		\label{cluster3s}
        \end{subfigure}  
             \begin{subfigure}{0.15\textwidth}
                \centering
        		\includegraphics[width=\textwidth]{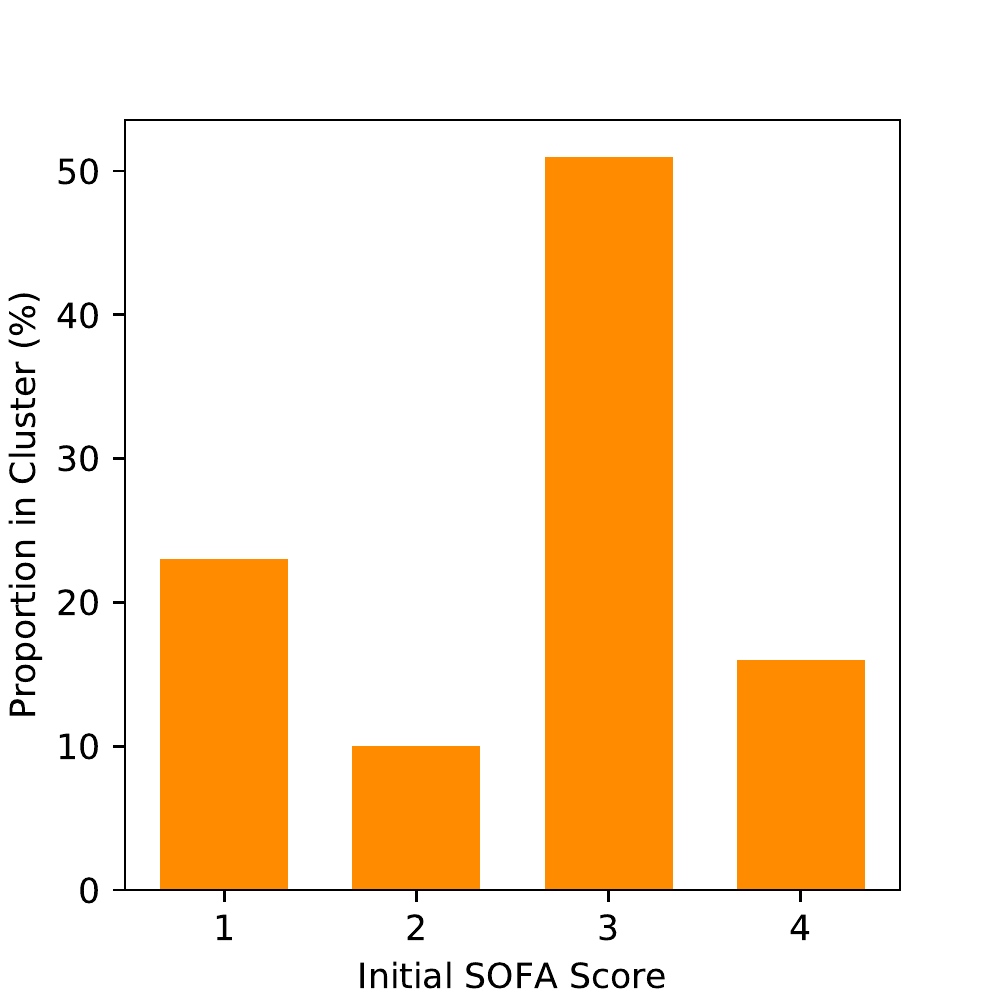}
        		\caption{SCE: 2.1 }
        		\label{cluster4s}
		\end{subfigure}  
		  \begin{subfigure}{0.15\textwidth}
                \centering
        		\includegraphics[width=\textwidth]{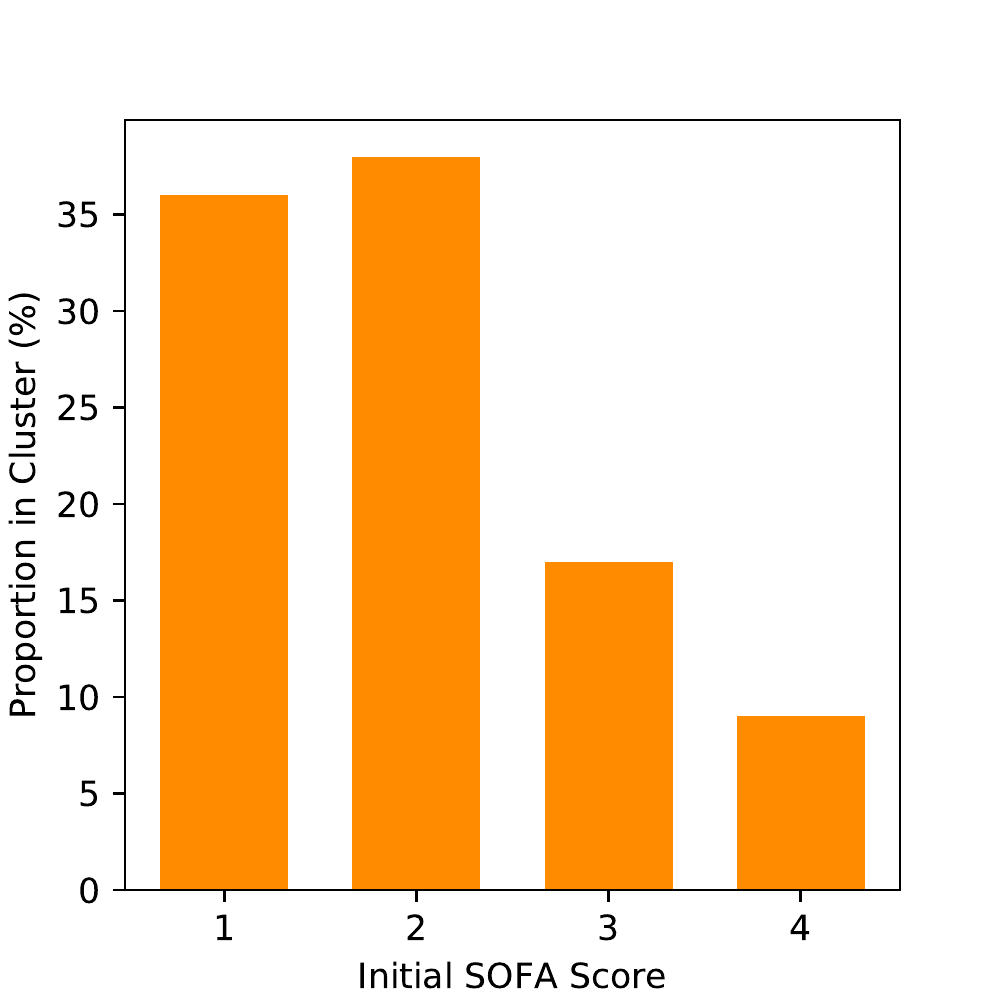}
        		\caption{SCE: -3.8 }
        		\label{cluster5s}
		\end{subfigure}  
		  \begin{subfigure}{0.15\textwidth}
                \centering
        		\includegraphics[width=\textwidth]{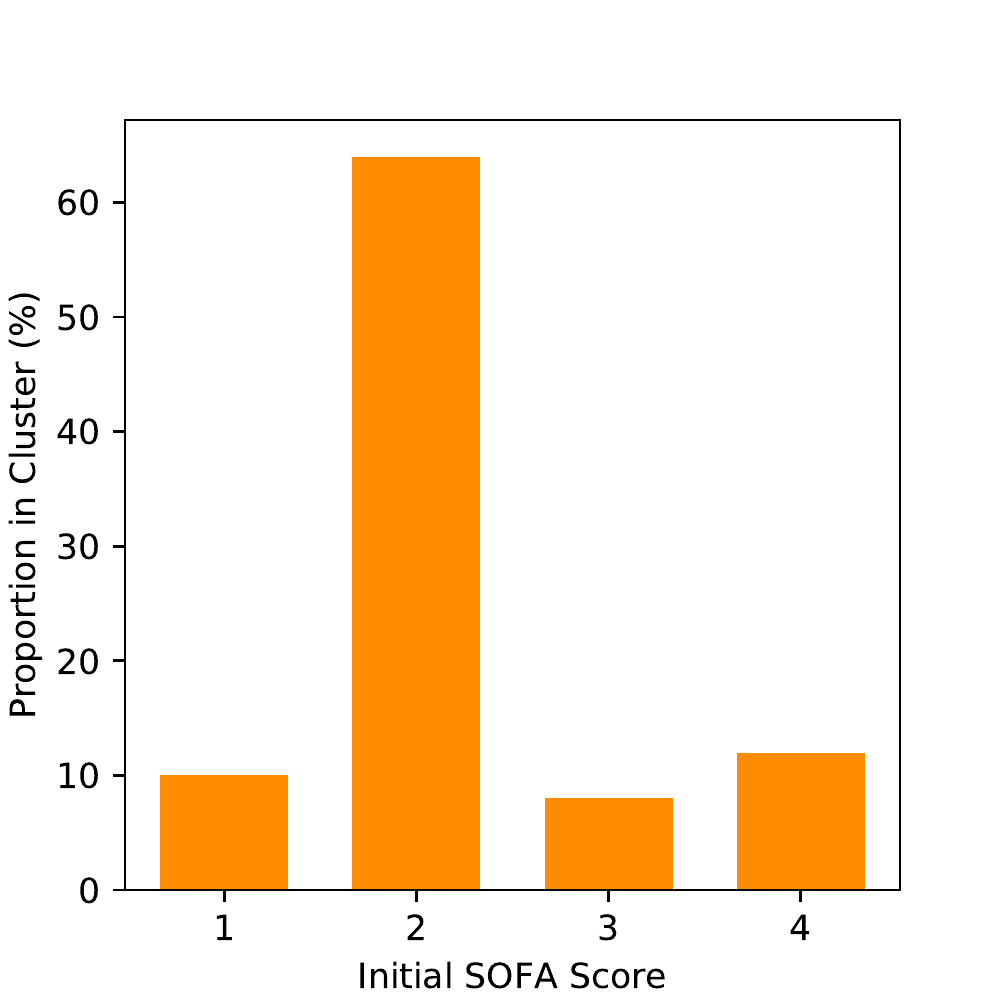}
        		\caption{SCE: 1.7 }
        		\label{cluster6s}
		\end{subfigure}  
		\caption{\label{sepsis clusters} Proportion of initial SOFA scores in each cluster. The variation in initial SOFA scores across clusters suggests that it is a potential confounder of odds of mortality when managing and treating sepsis.} 
\label{fig:sepsis-clusters}
\end{figure*} 
The information curves for $I(Z; T)$ and $I(Z; (Y, T))$ can be found in Figures \ref{information curve:c} and \ref{information curve:d} respectively. Overall, we  we can perform a sufficient reduction of the high-dimensional covariate information to between 4 and 6 dimensions, while accurately estimating $Y$. Since there is no ground truth available for the sepsis task, we do not have access to the true confounding variables. However, we analyse the clusters with respect to a patient's initial Sequential Organ Failure Assessment (SOFA) scores, used to track a patient's stay in hospital (Figure \ref{sepsis clusters}). Clusters 2, 5 and 6 tend to have higher proportions of patients with lower SOFA scores compared to Clusters 3 and 4. The SCE values over these clusters also vary considerably, suggesting that initial SOFA score is potentially a confounder of treatments and odds of in-hospital mortality. This is consistent with medical studies such as \cite{medam17, studnek2012impact} in which high initial SOFA scores tend to impact treatments and overall chances of survival. Finally, while we cannot quantify an error in estimating the ACE since we do not have access to the counterfactual outcomes, we can still compute the ACE for the sepsis management task. Here, we specifically observe a \emph{negative} ACE, suggesting that treating patients with vasopressors generally reduces the chances of mortality. Performing such analyses for tasks like sepsis may help adjust for confounding and assist in establishing potential guidelines.

%% file: conclusion.tex
\section{Conclusion}
\label{conclusion}
We have presented a novel approach to estimate causal effects with incomplete covariates at test time. This is a crucial problem in several domains such as healthcare, where doctors frequently have access to routine measurements, but may have difficulty acquiring for instance, genotyping data for patients at test time as a result of the costs. We used the IB framework to learn a sufficient statistic of information relevant for predicting outcomes. By further introducing a discrete latent space, we could learn equivalence classes that, in turn, allowed us to transfer knowledge from instances with complete covariates to instances with incomplete covariates, such that treatment effects could be accurately estimated. Our extensive experiments show that our method outperforms state-of-the-art approaches on synthetic and real world datasets. Since handling systematic missingness is a highly relevant problem in healthcare, we view this as step towards improving these systems on a larger scale.